\documentclass{article}

\PassOptionsToPackage{numbers, compress}{natbib}

     \usepackage[final]{neurips_2021}

\usepackage[utf8]{inputenc} 
\usepackage[T1]{fontenc}    
\usepackage{hyperref}       
\usepackage{url}            
\usepackage{booktabs}       
\usepackage{amsfonts}       
\usepackage{nicefrac}       
\usepackage{microtype}      
\usepackage{xcolor}         
\usepackage{authblk}

\usepackage{todonotes}
\usepackage{changes}

\usepackage[acronym]{glossaries}
\newacronym{smi}{SMI}{Soil Moisture Index}
\newacronym{spi}{SPI}{Standardized Precipitation Index}
\newacronym{spei}{SPEI}{Standardized Precipitation Evapotranspiration Index}

\glsdisablehyper

\title{
On the Generalization of Agricultural Drought Classification from Climate Data
}

\author[1,2]{Julia Gottfriedsen}
\author[2]{Max Berrendorf}
\author[3,4]{Pierre Gentine}
\author[1]{Birgit Hassler}
\author[5,6]{Markus Reichstein}
\author[7,1]{Katja Weigel}
\author[1,7]{Veronika Eyring}

\bigskip

\affil[1]{Deutsches Zentrum für Luft- und Raumfahrt (DLR), Institut für Physik der Atmosphäre, Oberpfaffenhofen, Germany}
\affil[2]{Ludwig-Maximilians-Universität München, Munich, Germany}
\affil[3]{Department of Earth and Environmental Engineering, Columbia University, NY, USA}
\affil[4]{Center for Learning the Earth with Artificial intelligence and Physics (LEAP), Columbia University, NY, USA}
\affil[5]{Department of Biogeochemical Integration, Max Planck Institute for Biogeochemistry, Jena, Germany}
\affil[6]{Michael‐Stifel‐Center Jena for Data‐driven and Simulation Science, Jena, Germany}
\affil[7]{University of Bremen, Institute of Environmental Physics (IUP), Bremen, Germany}

\begin{document}

\maketitle

\begin{abstract}
Climate change is expected to increase the likelihood of drought events, with severe implications for food security.
Unlike other natural disasters, droughts have a slow onset and depend on various external factors, making drought detection in climate data difficult. 
In contrast to existing works that rely on simple relative drought indices as ground-truth data, we build upon soil moisture index (SMI) obtained from a hydrological model.
This index is directly related to insufficiently available water to vegetation.
Given ERA5-Land climate input data of six months with landuse information from MODIS satellite observation, we compare different models with and without sequential inductive bias in classifying droughts based on SMI.
We use PR-AUC as the evaluation measure to account for the class imbalance and obtain promising results despite a challenging time-based split.
We further show in an ablation study that the models retain their predictive capabilities given input data of coarser resolutions, as frequently encountered in climate models.
\end{abstract}

\section{Introduction}
\label{Intro}

Drought is one of the most widespread and frequent natural disasters in the world, with profound economic, social, and environmental impacts~\citep{keyantash2002quantification}.
Unlike other natural hazards, droughts are a gradual process, often have a long duration, cumulative impacts, and widespread extent~\citep{below2007documenting}.
Climate change is expected to increase the area and population affected by soil moisture droughts and also the probability of extreme drought events comparable to the one of 2003 across Europe~\citep{IPCC2021, samaniego2018droughteurope}. 
Therefore, it is a critical scientific task to understand better possible changes in drought frequency and intensity under varying climate scenarios~\citep{King2020}. 
Drought is commonly classified into four categories: meteorological, agricultural, socioeconomic, and hydrological.
In this study, we focus on agricultural drought since it has a considerable impact on human population evolution~\citep{LloydHughes2013}.
Agricultural droughts can be quantified as a ``deficit of soil moisture relative to its seasonal climatology at a location''~\citep{sheffield2007characteristics}.
A low \gls{smi} in the root zone is a direct indicator of agricultural drought and inhibits vegetative growth, directly affecting crop yield and therefore food security~\citep{keyantash2002quantification}. 
The physical processes involved in drought depend on complicated interactions among multiple variables and are spatiotemporally highly variable.
This behavior makes droughts hard to predict, classify, and understand~\citep{below2007documenting}.
However, recently, machine learning (ML)-based methods have demonstrated their ability to capture hydrological phenomena well, e.g., rainfall-runoff~\citep{kratzert2018rainfall} and flood~\citep{le2019application}.
ML has also been applied to drought detection but relied on relative indices as labels due to the lack of ground truth data~\citep{BELAYNEH201637, Shamshirband2020, FENG2019303}.
Using such statistically derived labels can lead to unreliable detection of droughts in climate model projections and, accordingly, an inaccurate estimation of the impacts of future climate change~\citep{VicenteSerrano2010}. 
Therefore, we compare several ML algorithms in their ability to classify droughts based on agriculturally highly relevant soil moisture.
A future goal is to provide an ML-based drought classification for climate projections under various scenarios.
While we do not yet operate on climate model output from the Coupled Model Intercomparison Project (CMIP6)~\cite{Eyring2016}, 
in this work, we nevertheless showcase that drought classification is possible with the variables available in the output of CMIP6 climate projections, and thus it is promising to further pursue the goal.

\section{Data Preparation}

\begin{figure}
    \centering
    \includegraphics[
        height=3.7cm, keepaspectratio
    ]{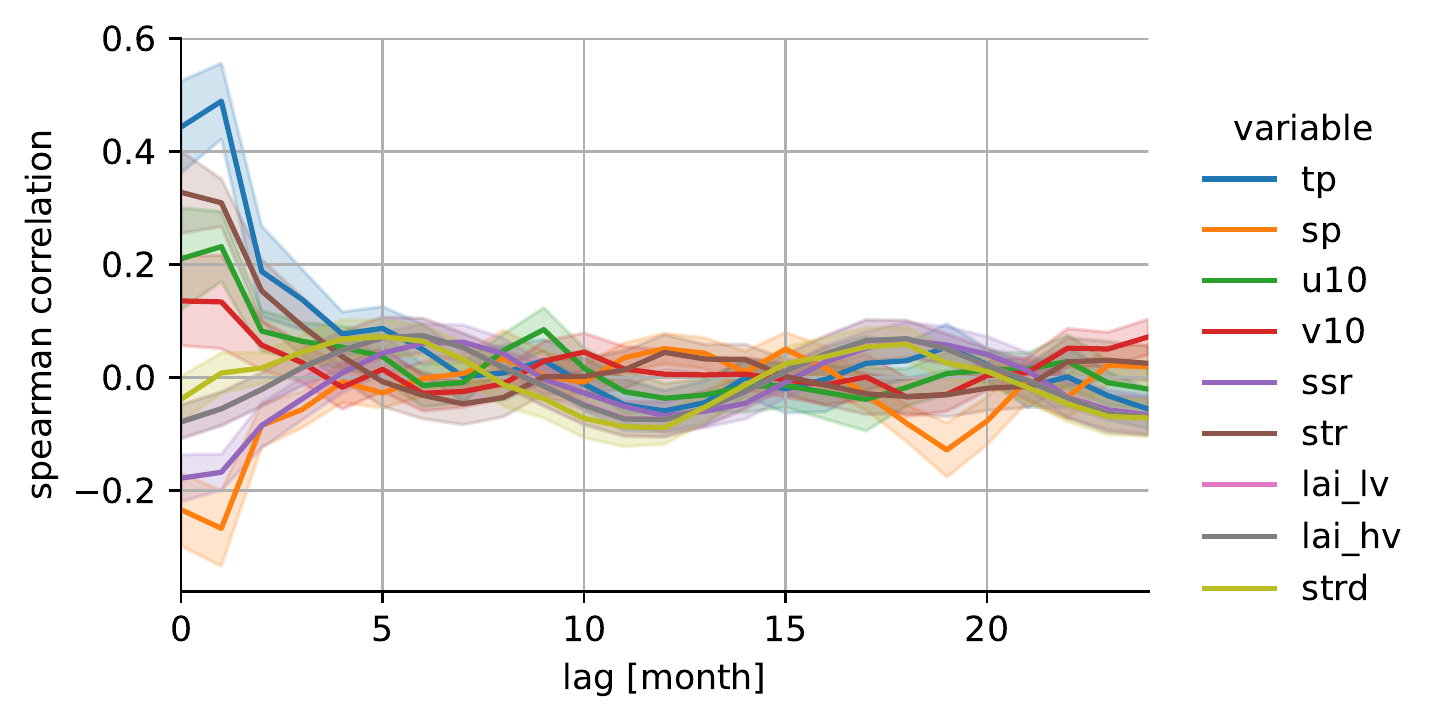}
        \includegraphics[
        height=4.5cm, 
        keepaspectratio
    ]{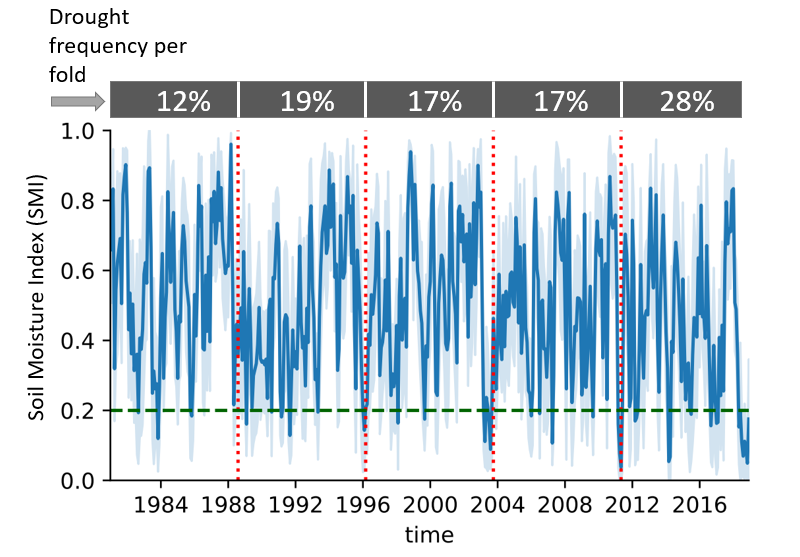}
    \caption{
        {\it Left:} Time-lagged Spearman correlation between the selected ERA5-Land input variables and the target variable SMI over 24 months. {\it Right:} Time series of \gls{smi} from 1981-2018 from the Helmholtz dataset. 
        The shaded area shows the standard deviation across different locations.
        Red dotted lines show the split points for the time-based split into $k$ folds, and the green dashed line shows the binarization threshold for drought events. Shown above is the frequency of the positive class (drought events) per fold.
    }
    \label{fig:lagged-input-correlation_and_class-imbalance}
\end{figure}

Low soil moisture levels depend on various meteorological input variables and the soil type. Retrieving accurate SMI ground-truth data is therefore complicated:
Spatially-continuous soil moisture data on a resolution smaller than 0.25 degree is only available from satellite observations or model simulations. 
Satellite observations are exclusively available for recent years, include only the top few centimeters of the soil, and have data gaps due to unfavorable data retrieval conditions such as snow or dense vegetation~\citep{Dorigo2017}.
Therefore, we select modeled SMI data as the ground-truth label.
Due to SMI data availability, the selected experiment region is Germany.
The data is limited to January 1981 to December 2018 by the availability of an overlapping period from both ERA5-Land and the SMI data. 
All datasets used in this study are freely available.

The target variable SMI is derived from the German drought monitor uppermost 25cm of soil data as \gls{smi} labels~\citep{zink2016german}, which is generated by the hydrological model system based on data from about 2,500 weather stations~\cite{Samaniego2010, Kumar2013}.
Figure~\ref{fig:lagged-input-correlation_and_class-imbalance} shows the SMI distribution over time and the chosen binarization threshold. 

We use monthly time-series of 12 selected variables from the ERA5-Land reanalysis, e.g., pressure, precipitation, temperature (see Table~\ref{tab:era5-variables}). 
We selected ERA5-Land due to its higher resolution compared to ERA5 (9 km vs. 31 km) and its consequently better suitability for land applications~\citep{MuozSabater2021}.
To isolate the causal effects on SMI and avoid short-cut learning, we do not include potential confounding factors such as evaporation, runoff, and skin temperature.
We also deliberately restrict the input variables to those commonly available in the latest generation of climate models to enable the transfer of the trained models to data directly obtained from climate model simulations.

Land use and vegetation type data based on the MODIS (MCD12Q1) Land Cover Data is used as an input feature, represented as soil type fractions~\citep{friedl__mark_mcd12q1_2019}.


\textbf{Interpolation and Label Derivation} 
Drought is an extreme weather event. Extreme events occur at the tail of variable distributions.
Thus, we chose a classification setting with the tail of the SMI distribution as labels instead of a regression setting.
The input data is re-gridded to the ERA5-Land regular latitude-longitude grid ($0.1^\circ \times 0.1^\circ \approx (9km)^2$).
In this paper, we follow the drought classification from the German and U.S. drought monitors~\citep{svoboda2002drought, zink2016german} using an \gls{smi} threshold of $0.2$.


\textbf{Dataset Split} 
As seen in Figure~\ref{fig:lagged-input-correlation_and_class-imbalance}, the \gls{smi} values for the same location exhibit a noticeable but declining correlation for lags up to 6 month.
A simple random split over data points could therefore lead to data leakage, where memorizing \gls{smi} values from training and simple interpolation can lead to erroneously good results.
Thus, we opt for a modified $k$-fold time-series split.
First, we evenly determine $k-1$ split times to create $k$ time intervals.
For the $k$th split, we train on folds $\{1, \ldots, k\}$, validate on fold $k+1$ and test on $k+2$.
This split enables us to better assess parameter stability over time, mimicking increasing climate projection length. 
We decide to use $k=5$ as a good compromise between a sufficient number of folds for a robust performance estimate and large enough folds with multiple years of data to account for seasonal and interannual effects.
Figure~\ref{fig:lagged-input-correlation_and_class-imbalance} shows the resulting folds separated by red dotted lines.
Note that the drought sample availability (positive class) varies between folds from 12\% to 28\%.


\section{Methodology}  \label{Methodology}

We frame drought classification as a binary classification problem given climate, land use as well as location data.
Since the \emph{memory-effect}~\citep{Kingtse2011} is suspected of playing an essential role in the development of droughts, we frame the problem as a 
\emph{sequence classification}: The models use a window of the last $k$ months of climate input data for the current location to predict the drought label at the current time step.
In addition to the climate variables, we also provide a positional and seasonal encoding as input features:
For the positional encoding, we directly use the latitude \& longitude grid values.
A 2D circular encoding considers the seasonality based on the month of the year (\textit{month}).
$
s = \left[
\cos \left(2 \pi \cdot \frac{\textit{month}}{12}\right);
\sin \left(2 \pi \cdot \frac{\textit{month}}{12}\right)
\right]
$, 
where $[\cdot;\cdot]$ denotes the concatenation.
Besides using the location as an input feature, we do not explicitly include inductive biases for spatial correlation.
Due to missing values and a non-rectangular shape of the available data area, simple grid-based methods such as a 2D-CNN are not directly applicable.
The exploration of methods for irregular spatial data, such as those described in~\citep{DBLP:journals/spm/BronsteinBLSV17}, will be a focus of future work.

\textbf{Addressing Imbalanced Data}
In the entire dataset, examples for the drought class account for 18\% of the total samples.
We address this class imbalance by adding class weights proportional to the inverse class frequency during training and using an appropriate metric, PR-AUC, during evaluation.

\textbf{Input Sequence Length}
The determination of a suitable sequence length is based on the Spearman correlation of the climate variables and the target \gls{smi} variable and the lagged correlation of the \gls{smi} variable, as both shown in Figure~\ref{fig:lagged-input-correlation_and_class-imbalance}. Cyclical and non-cyclical decaying dependencies are considered, and both are indeed observed. 
Therefore, we select a window size of six months for our models, which in line with the period commonly used on monthly mean data by other drought indices such as the \gls{spi}~\citep{mckee1993relationship}, and the \gls{spei}~\citep{VicenteSerrano2010}.

\textbf{Models} \label{Models}
We investigate support vector machines (SVMs) (\textbf{M1}) with linear kernels as well as an MLP model which receives the flattened window as a single large vector as input (\textbf{M2}), which we denote by \texttt{dense}.
To investigate whether an explicit inductive bias for sequential data is beneficial, we also include two main sequence encoders to obtain a representation of the input sequence for the sequence prediction.
The \texttt{cnn} model (\textbf{M3}) applies multiple 1D convolutional layers before aggregating the input sequence to a single vector representation by average pooling.
The \texttt{lstm} model (\textbf{M4}) uses multiple LSTM layers and the final hidden state as the sequence representation.
For both sequence encoders, the drought classification is obtained by a fully connected layer on top of this representation.

\textbf{Experimental Setup} 
We use \texttt{sklearn}~\citep{pedregosa2011scikit} for the 
SVM, and implement the other models in \texttt{tensorflow}~\citep{ https://doi.org/10.5281/zenodo.4758419}.
To reflect the considerable class imbalance, we choose the 
area under the precision-recall curve (PR-AUC) as evaluation measure, which does not neglect the model's performance for the minority/positive class, i.e., droughts. 
For hyperparameter optimization we use \texttt{ray tune}~\citep{liaw2018tune} with random search instead of grid search due to its higher efficiency~\citep{bergstra12}.
The best hyperparameters are selected per fold according to validation PR-AUC on the fold's validation data, and we report test results of the corresponding model trained across five different random seeds.
The climate variables of the dataset are normalized to $[0, 1]$.

\section{Results and Discussion}

\begin{figure}
    \centering
    \includegraphics[height=3.4cm,keepaspectratio]{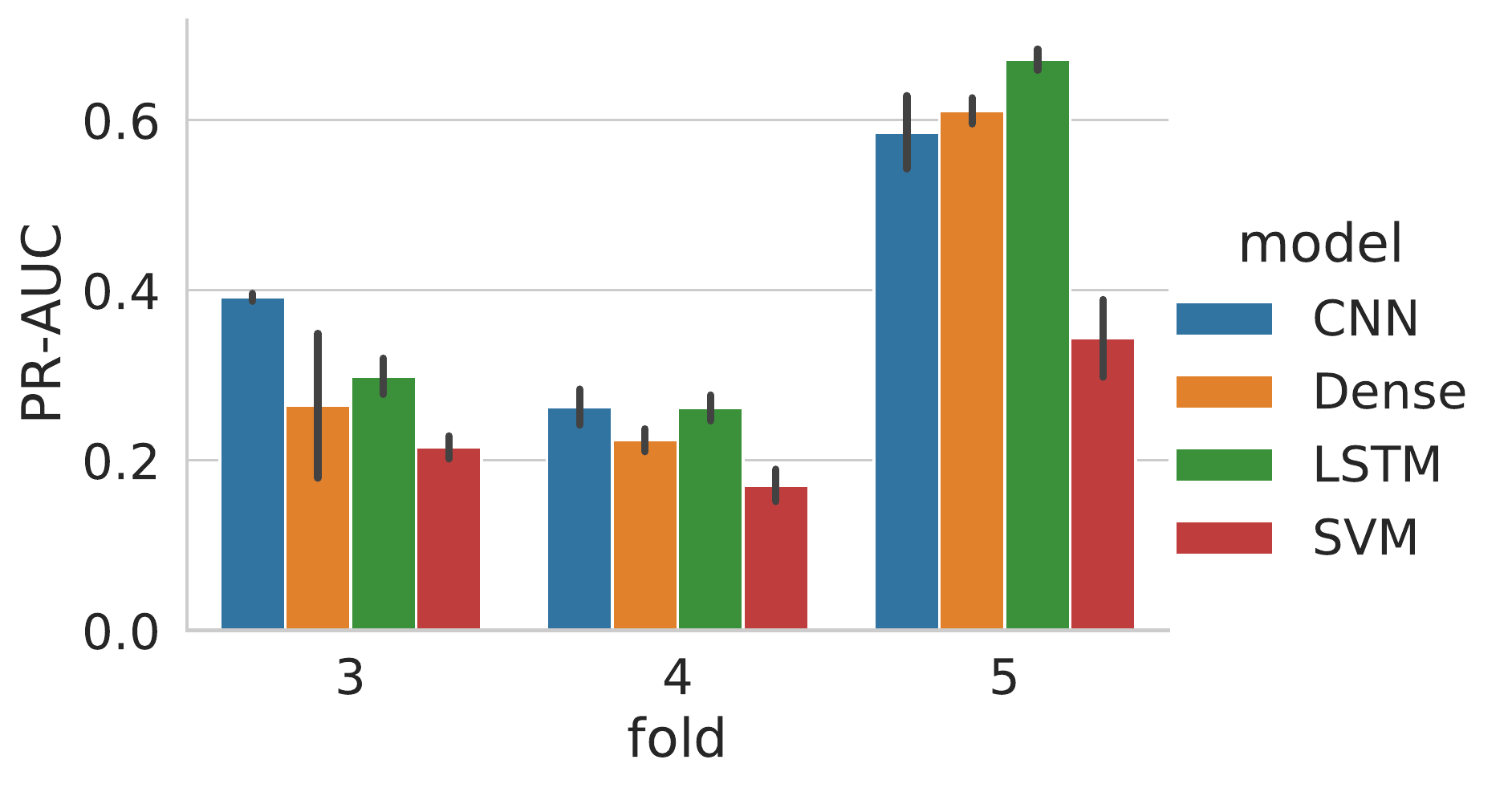}
    \includegraphics[height=3.4cm, keepaspectratio]{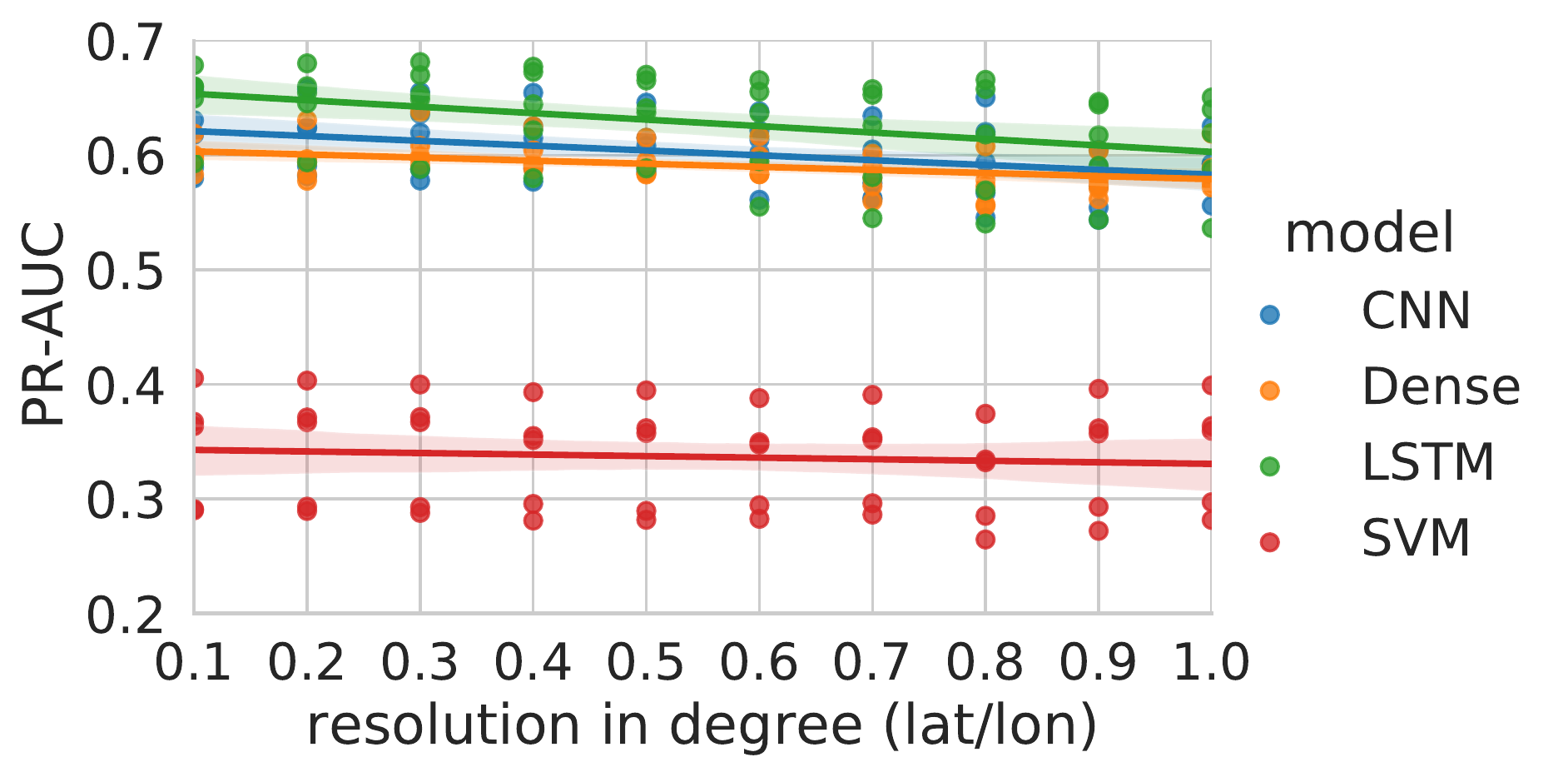}
    \caption{\textit{Left:} Results on PR-AUC of the different models on the test dataset across five different random seeds for drought classification using a window of six months. \textit{Right: } Ablation study:Inference on models trained on high resolution given input with decreasing resolution. Evaluation on five different random seeds using a window of six months.}
    \label{fig:results2}
\end{figure}

\textbf{Model Comparison}
The resulting architectures were selected based on the validation PR-AUC on the second fold to account for a large variety of drought causes in the training data. The resulting hyperparameters are listed in Table~\ref{hyperparams}.
The results are shown in Figure~\ref{fig:results2},   and Table~\ref{tab:results}.
We observe that the PR-AUC is larger than the class frequency of the positive class indicating that the models indeed learned a non-trivial relation between the input variables and the target.  The results for the F1 score can be found in Figure~\ref{fig:results1} with F1 scores larger than 0.5.
Moreover, the performance varies for different folds, highlighting the challenging setting of a time-based split, where distributions can differ between different folds.
There is no clear winner between the architectures: all models except the linear SVM perform comparable across folds.
In particular, we do not observe a significant difference between models with an explicit inductive bias for sequential data.
Since the utilized SMI data describes only the uppermost 25cm of the soil, the suspected memory effect might be more prominent in deeper soil layers. 
Our initial data analysis supports this, with the correlation of the input variables with the target being strongest close in time, cf. Figure~\ref{fig:lagged-input-correlation_and_class-imbalance}. 

\textbf{Ablation: Coarsening the Data Resolution}
As an important future application of our models is on simulated climate data from climate models, we investigate further how the performance is affected by changing the resolution from the original $0.1^\circ$ to a coarser spatial resolution.
The horizontal resolution of CMIP6 models varies from around 0.1$^{\circ}$ to 2$^{\circ}$ in the atmosphere~\cite{Cannon2020}.
Given the regional restriction of our input data, we restrict the ablation study to a range of 0.1$^{\circ}$-1.0$^{\circ}$ with 0.1$^{\circ}$ steps.
The architecture performing best on 0.1$^{\circ}$ is used in inference to calculate the results on the coarser resolutions without re-training.

On the right-hand side of Figure~\ref{fig:results2} we visualize the results of the resolution ablation.
In general, we observe a negative correlation between resolution and performance.
The LSTM architecture is most affected by this but also generally shows the noisiest results overall.
Overall, the models trained on 0.1$^{\circ}$ input data show satisfactory performance when applied to coarser input data without dedicated training.
This promising result indicates that it is possible to predict drought events under varying future climate scenarios with models trained on fine-grained drought labels.

\section{Summary and Outlook}
We summarize our contributions as follows: (1) We are the first to compare several ML models in their capability of classifying agricultural drought in a changing climate based on soil moisture index (SMI).
We use ground truth data from a hydrological model and intentionally restrict the climate input variables to those available in the newest generation of CMIP6 climate models. 
We also include land use information.
(2) We provide an ablation study regarding a transfer to coarser input data resolution, demonstrating that the model capabilities are transferable to lower resolution when trained in higher resolution. 

In future work, we plan to use climate model output as input data for our algorithm to produce drought estimates under varying future scenarios. 
This will facilitate the transfer from learning on real input data to input data obtained from simulations.
Apart from feeding the location information encoded as an additional input feature to the model, we plan to add location-aware models motivated by the strong regional correlation of the input variable as seen in Figure~\ref{fig:spatial-corr}. 
Additionally, we plan to investigate other ground truth labels, e.g., SMAP~\citep{smapL3} and expand the study region globally. 

Overall, we consider our study as an important step towards machine learning-based agricultural drought detection.
With our intentional restriction to variables available in climate models, we pave the way towards application on simulated data, thus facilitating the investigation of agricultural droughts in a changing climate.



\begin{ack}
The work for this study was funded by the European Research Council (ERC) Synergy Grant “Understanding and Modelling the Earth System with Machine Learning (USMILE)” under Grant Agreement No 855187. This manuscript contains modified Copernicus Climate Change Service Information (2021) with the following dataset being retrieved from the Climate Data Store: ERA5-Land (neither the European Commission nor ECMWF is responsible for any use that may be made of the Copernicus Information or Data it contains). Ulrich Weber from the Max Planck Institute for Biogeochemistry contributed pre-formatted MCD12Q1 MODIS data. SMI data were provided by the UFZ-Dürremonitor from the Helmholtz-Zentrum für Umweltforschung. The computational resources provided by the Deutsches Klimarechenzentrum (DKRZ, Germany) were essential for performing this analysis and are kindly acknowledged.

\end{ack}

\bibliography{main}
\bibliographystyle{icml2021}

\clearpage

\section*{Appendix}
\appendix


\renewcommand{\thefigure}{A\arabic{figure}}
\renewcommand{\thetable}{A\arabic{table}}
\setcounter{figure}{0}
\setcounter{table}{0}

\begin{figure}[h]
    \centering
    \includegraphics[
    width=\linewidth,
     height=13cm, keepaspectratio,
    ]{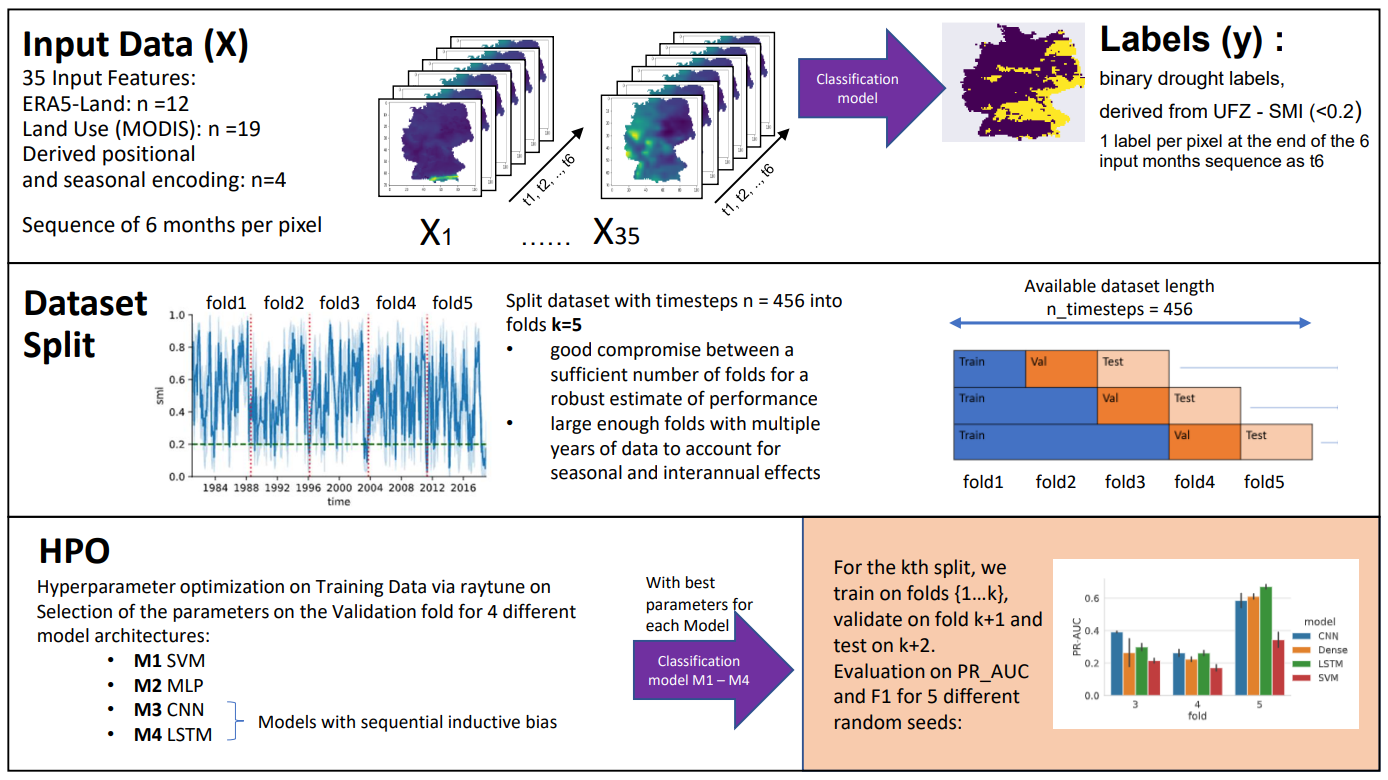}
    \caption{
    The first row displays the overall approach for classifying agricultural droughts by using a thresholded soil moisture index (SMI) from a hydrological model as ground truth labels per grid cell.
    Sequential input per location (6 months) are 35 input variables. 
    The middle row shows the data split.
    We use a modified $k$-fold time-series split.
    The lower row displays the training and evaluation {\it right} and the hyperparameter optimization {\it left} as described in Section~\ref{Methodology}.
    The best hyperparameters are selected on the validation data. Due to the high imbalance in the dataset, for the evaluation, we use the area under the precision-recall curve (PR-AUC) metric to also focus on the model’s performance in correctly identifying the minority/positive class (drought).
    The evaluation results are reported on the test part of a split across five different random seeds.
    }
    \label{fig:schematic}
\end{figure}

\begin{table}
    \centering
    \small
    \caption{Overview of the variables used in this study. Native resolution of SMI: 4x4l, ERA5-Land: 9km, MODIS land use: 500mx500m }
    \vskip 0.15in
    \label{tab:era5-variables}
    \begin{tabular}{l|l|p{8.8cm}|l}
        \toprule
        source & variable & description & unit \\
        \midrule
        Helmholtz & SMI  &  soil moisture index topsoil (top25cm) via UFZ Drought Monitor & - \\ 
        ERA5-Land & u10, v10 & wind (u + v component at 10m) & $m s^{-1}$  \\ 
           &  tp & total precipitation & $m$ \\
           & sp & surface pressure & $Pa$ \\
           &  t2m &  temperature  & $K$ \\
           &  ssrd & 	surface solar radiation downwards & $J m^{-2}$ \\ 
           &  d2m & dewpoint temperature  & $K$ \\
           & ssr & surface net solar radiation & $J m^{-2}$ \\ 
           & str & surface net thermal radiation & $J m^{-2}$ \\ 
           & lai\_lv, lai\_hv & leaf area index high + low vegetation & $m^2 m^{-2}$ \\
           &  strd & surface thermal radiation downwards &  $J m^{-2}$\\ 
        MODIS & land use class & water, evergreen needleleaf forest, Evergreen Broadleaf forest,  Deciduous Needleleaf forest, Deciduous Broadleaf forest, Mixed forest, Closed shrublands, Open shrublands, Woody savannas, Savannas, Grasslands, Permanent wetlands, Croplands, Urban and built up, Cropland Natural vegetation mosaic, Snow and ice, Barren or sparsely vegetated, Cropland &  Fraction \\
        self-derived & positional encoding & latitude longitude grid & degree \\
        self-derived & seasonal encoding &  2D circular encoding of the month & degree \\ 
        \bottomrule
    \end{tabular}
\end{table}

\begin{table}
\caption{
Soil condition classification based on \gls{smi} according to~\citep{svoboda2002drought, zink2016german}.
}
\label{drought-classes}
\begin{center}
\begin{tabular}{ll}
\toprule
\gls{smi} &   soil condition  \\
\midrule
$(0.2,\phantom{0} 0.3\phantom{0}]$ & abnormally dry  \\
$(0.1,\phantom{0} 0.2\phantom{0}]$ & moderate drought \\
$(0.05, 0.1\phantom{0}]$ & severe drought \\
$(0.02, 0.05]$ & extreme drought \\
$(0,\phantom{.00} 0.02]$ & exceptional drought \\
\bottomrule
\end{tabular}
\end{center}
\end{table}


\begin{figure}
    \centering
    \includegraphics[height=5cm,keepaspectratio]{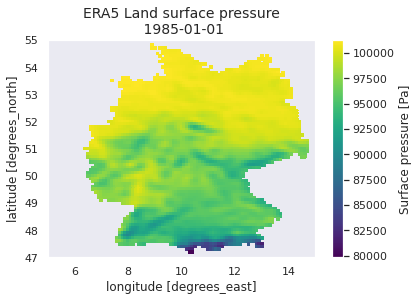}
    \includegraphics[height=5cm,keepaspectratio]{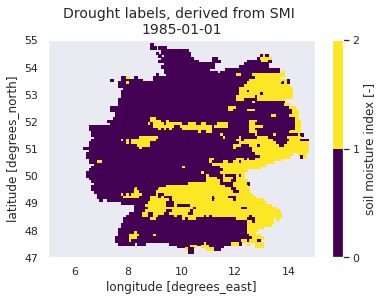}
    \caption{Data examples for 1 month. {\textit{Left:} ERA5 input variable example "pressure" \textit{Right:} Target variable: Binarized SMI}}
    \label{fig:data_example}
\end{figure}

\begin{table*}
    \centering
    \small
    \caption{The Hyperparameters resulting from the random search for binary drought classification.}
    \label{hyperparams}
    \label{tab:results}
    \begin{tabular}{lrrrrrrr}
        \toprule
        type & HPO fold & hidden & lr & dropout & activation & batchnorm & batch size\\
        \midrule
        LSTM    & 2 &  16, 32 & $1.18\text{e-}4$& 0.1 & softplus  & False  & 2208 \\   
                & 3 & 96, 96   & $1.00\text{e-}4$& 0.2  & relu & False  & 96 \\   
                & 4 &  32, 48, 128  & $2.15\text{e-}5$ & 0.0  & softplus  & True  & 2592\\
        \midrule
        CNN     & 2 &  128, 176, 224, 240  & $3.53\text{e-}5$ & 0.1  & softplus  & False  & 32\\   
                & 3 & 112, 176  & $2.40\text{e-}5$ & 0.2  & softplus  & False  & 64 \\ 
                & 4 & 16, 96, 128  & $1.29\text{e-}2$ & 0.1 & ReLu  & True  & 448\\
        \midrule
        Dense   & 2 & 32, 48, 96  & $3.36\text{e-}2$ & 0.1 & relu  & False  & 769 \\   
                & 3 & 48, 208, 208, 208 & $1.66\text{e-}2$ & 0.2 & softplus & True  & 192  \\  
                & 4 & 80, 192  & $1.02\text{e-}5$ & 0.2  & ReLu  & True  & 800\\
        \bottomrule
    \end{tabular}
\end{table*}

\begin{table*}[t]
    \centering
    \caption{
    Results on the test dataset across five different random seeds for drought classification using a window of six months.
    }
    \label{tab:results}
    \vskip 0.15in
    \begin{tabular}{llrrrrrr}
        \toprule
         &   & \multicolumn{2}{c}{Macro F1} & \multicolumn{2}{c}{PR-AUC} \\
         &   &                      mean &       std &           mean &       std              \\
        model & test fold &                           &           &                &           \\
        \midrule
        dense & 3 &                      0.5340 &  0.0724 &         0.2640 &  0.0950  \\
              & 4 &                      \textbf{0.5212} &  0.0359 &         0.2234 &  0.0148   \\
              & 5 &                     0.6424 &  0.0272 &         0.6106 &  0.0170  \\
        \midrule
        lstm & 3 &                    0.5722 &  0.0176 &         0.2986 &  0.0236  \\
             & 4 &                    0.5096 &  0.0494 &         0.2614 &  0.0168 \\
             & 5 &                    \textbf{0.6648} &  0.0311 &         \textbf{0.6708} &  0.0138 \\
        \midrule
        cnn & 3 &                    \textbf{0.5976} &  0.0101 &         \textbf{0.3914} &  0.0050 \\
            & 4 &                    0.4766 &  0.0189 &         \textbf{0.2624} &  0.0235 \\
            & 5 &                    0.5650 &  0.0520 &         0.5854 &  0.0480\\
        \midrule
        SVM & 3 &                    0.5642 &  0.0352 &         0.2150 &  0.0150 \\
                  & 4 &                     0.4858 &  0.0154 &         0.1704 &  0.0210  \\
                  & 5 &                    0.4772 &  0.1437 &         0.3432 &  0.0508 \\
        \bottomrule
    \end{tabular}
\end{table*}

\begin{figure}
    \centering
    \includegraphics[height=3.4cm,keepaspectratio]{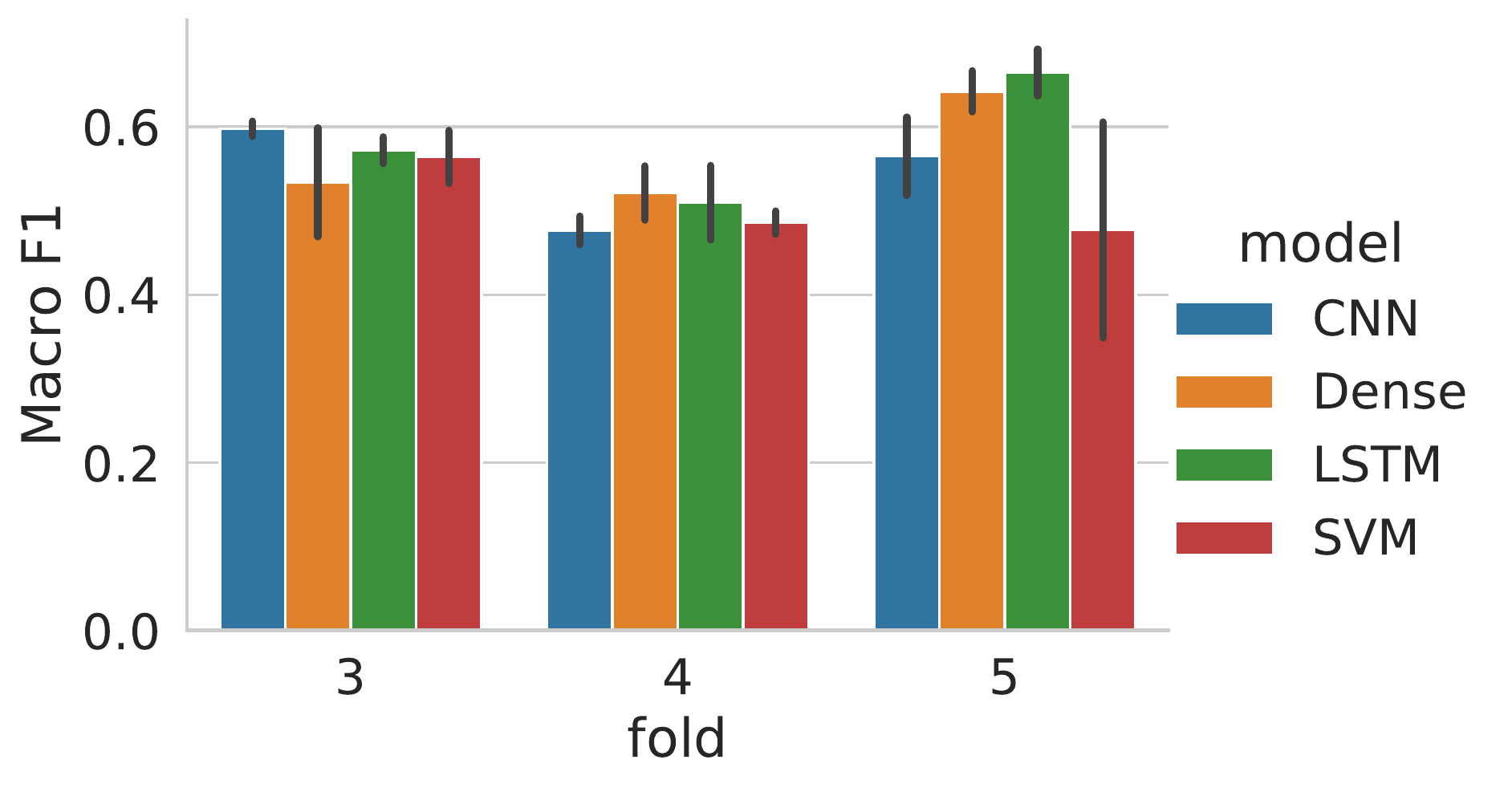}
    \includegraphics[height=3.4cm,keepaspectratio]{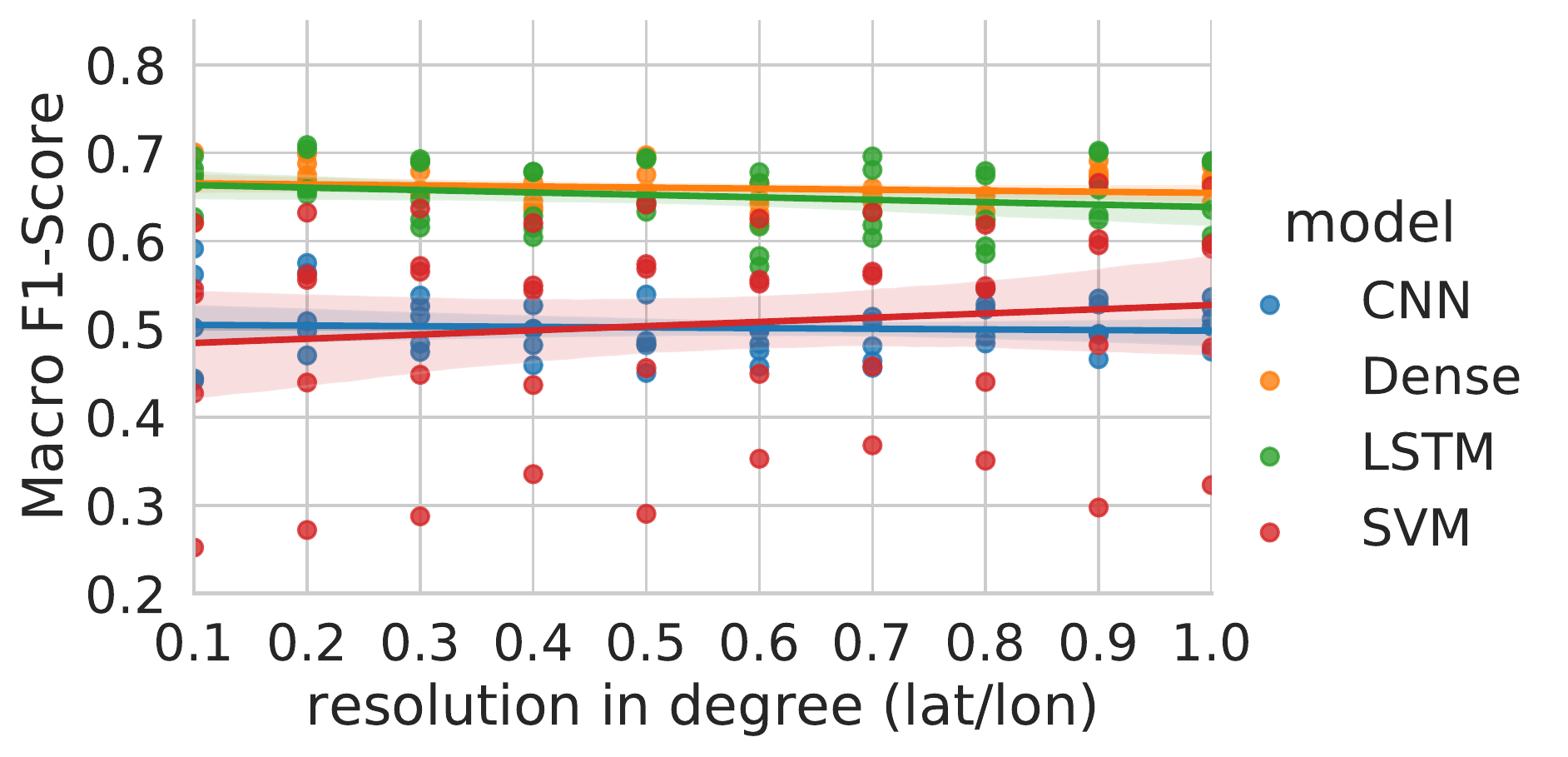}
    \caption{\textit{Left:} Results on F1 Score of the different models on the test dataset across five different random seeds for drought classification using a window of six months. \textit{Right:} Ablation study: Inference on models trained on high resolution given input with decreasing resolution. Evaluation on five different random seeds using a window of six months.}
    \label{fig:results1}
\end{figure}

\begin{figure}
    \centering
    \includegraphics[
        height=6cm, keepaspectratio,
    ]{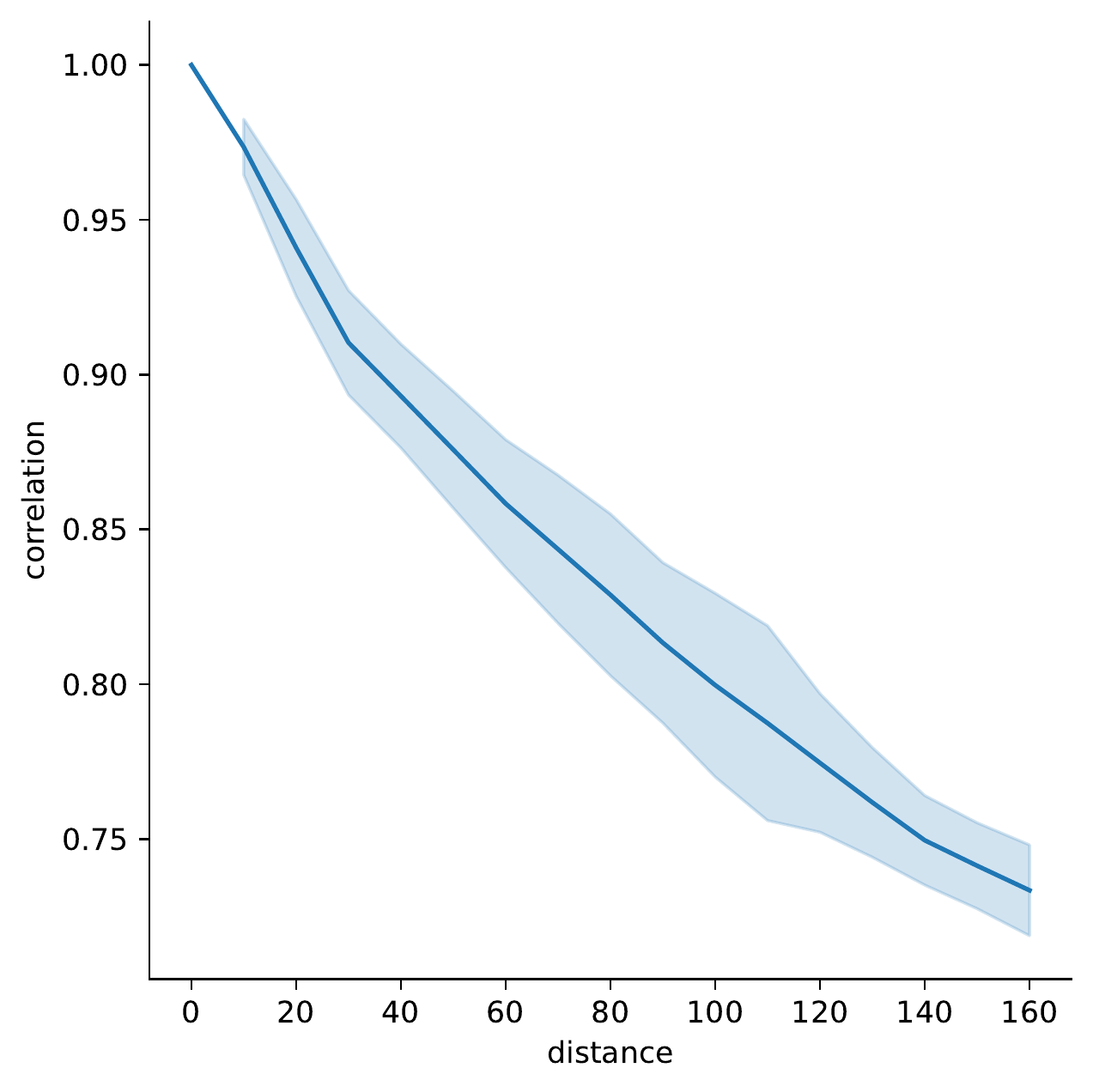}
    \caption{
        Time-lagged Spearman autocorrelation for the SMI target variable at the same location.
        The shaded area shows the standard deviation of the Spearman correlation across all analyzed locations. 
    }
    \label{fig:spatial-corr}
\end{figure}

\end{document}